\pdfoutput=1

\documentclass[11pt]{article}

\usepackage{EMNLP2023}
\usepackage{times}
\usepackage{latexsym}
\usepackage[T1]{fontenc}
\usepackage[utf8]{inputenc}
\usepackage{microtype}
\usepackage{inconsolata}
\usepackage{lipsum}
\usepackage{graphicx}
\usepackage{covington}
\usepackage{hwemoji}

\title{Emergent Inabilities? Inverse Scaling Over the Course of Pretraining}

\author{James A. Michaelov \and Benjamin K. Bergen \\
        Department of Cognitive Science, University of California San Diego \\ 
  \texttt{\{j1michae,bkbergen\}@ucsd.edu} }

\begin{document}
\maketitle
\begin{abstract}
Does inverse scaling only occur as a function of model size, or can it also occur over the course of training? We carry out an exploratory study investigating whether the performance of language models on specific tasks can decrease (while general performance remains high) during training on the language modeling task. We find 8 tasks on which Pythia 12B \citep{biderman_2023_PythiaSuiteAnalyzing} shows decreased performance over the course of training. Five of these tasks (\textsc{TruthfulQA-MC1}, \textsc{TruthfulQA-MC2}, \textsc{Hindsight Neglect}, \textsc{Memo Trap}, and \textsc{Pattern Match Suppression}) additionally show a consistent relationship whereby larger language models show a greater decrease in performance the more they are trained, despite showing standard (positive) scaling overall. This highlights the importance of testing performance at all relevant benchmarks any time models are trained on additional data, even if their overall performance improves.
\end{abstract}

\section{Introduction}
\label{sec:intro}
For language models, bigger is usually better. Recent research has found that both increased number of model parameters and increased size of the training dataset positively influence model performance \citep{brown_2020_LanguageModelsAre,kaplan_2020_ScalingLawsNeural,chowdhery_2022_PaLMScalingLanguage,clark_2022_UnifiedScalingLaws,du_2022_GLaMEfficientScaling,rae_2022_ScalingLanguageModels,hoffmann_2022_EmpiricalAnalysisComputeoptimal,thoppilan_2022_LaMDALanguageModels,wei_2022_EmergentAbilitiesLarge,taylor_2022_GalacticaLargeLanguage,srivastava_2022_ImitationGameQuantifying,touvron_2023_LLaMAOpenEfficient}. One particularly striking pattern that has been reported is \textit{emergence}, a nonlinearity in these relationships, where at a particular scale, language models improve rapidly at a given task \citep{wei_2022_EmergentAbilitiesLarge}.

However, while increased scale usually leads to improved performance, on certain tasks it correlates with decreased performance. This is known as \textit{inverse scaling} \citep{lin_2022_TruthfulQAMeasuringHow}. An example of a task on which inverse scaling is observed is the TruthfulQA benchmark, where larger language models are more likely to predict popular misconceptions over statements of fact \citep{lin_2022_TruthfulQAMeasuringHow}. More recently, additional tasks that reportedly show such an effect have been identified as part of the Inverse Scaling Prize \citep{mckenzie_2023_InverseScalingWhen}, as well as by other researchers \citep{jang_2023_CanLargeLanguage,michaelov_2023_RarelyProblemLanguage}.

Inverse scaling is a serious concern for several reasons. At a high level, inverse scaling may indicate `outer misalignment' \citep{perez_2022_AnnouncingInverseScaling} between the model training approach and the purposes to which they are applied. The lack of robustness observed in inverse scaling phenomena may thus indicate that the apparent successes of specific language models at a wide range of benchmarks \citep[e.g.,][]{hendrycks_2021_MeasuringMassiveMultitask,srivastava_2022_ImitationGameQuantifying} do not necessarily entail that they have the capability ostensibly being tested \citep{bowman_2021_WhatWillIt,raji_2021_AIEverythingWhole}.

The existence of inverse scaling is also concerning because of the possibility of other as yet unidentified tasks where performance similarly scales inversely with model size. Models that perform well on a variety of tasks may well present deteriorating performance in unseen tasks with scale, even as performance at established benchmarks increases. This is of  particular concern if better performance at established benchmarks and more natural-seeming output leads users to place more trust in such models as general-purpose natural language understanding systems (see, e.g., \citealp{bender_2021_DangersStochasticParrots}, for general discussion of such risks).

Finally, inverse scaling is also of concern because it is often unpredictable. In the same way that certain capabilities appear to emerge at scale \citep{wei_2022_EmergentAbilitiesLarge}, inverse scaling also appears or accelerates at given scales. For example, as \citet{mckenzie2022round1} show, the performance of Gopher \citep{rae_2022_ScalingLanguageModels} and Plain LM \citep{ganguli_2022_RedTeamingLanguage} at the Inverse Scaling Prize's negated question-answering task (\textsc{NeQA}) appears to be stable or even increasing as model size increases, only dropping as model size increases to around 7 billion parameters and beyond \citep{mckenzie2022round1}. Thus, inverse scaling may occur not just for unidentified tasks, but also for well-established tasks: a model architecture that performs well at a benchmark at a small scale may suddenly perform surprisingly worse as scale increases--it is not safe to assume that performance will continue to improve or even remain stable.

While previous work has focused on inverse scaling based on the number of model parameters \citep{lin_2022_TruthfulQAMeasuringHow,mckenzie2022inverse,mckenzie2022round1,mckenzie2022round2,jang_2023_CanLargeLanguage,michaelov_2023_RarelyProblemLanguage}; as discussed, scaling effects more generally  occur not just in relation to model size but also as a function of training data quantity. Recent work has shown that this latter effect has been substantially underestimated \citep{hoffmann_2022_EmpiricalAnalysisComputeoptimal,touvron_2023_LLaMAOpenEfficient}. Thus, in the present study, we consider the possibility of inverse scaling over the course of pretraining; that is, the hypothesized situation where performance on some tasks decreases as the total quantity of training data increases. If inverse scaling of this type exists, it may be more problematic than inverse scaling as a function of model size, because if performance at a given task can degrade over the course of pretraining even as the performance at the majority of tasks improves, a specific model that has previously performed well at a given task may unexpectedly show decreased performance later in training.

We test how well the 8 models of the Pythia \citep{biderman_2023_PythiaSuiteAnalyzing} suite perform at 12 tasks over the course of training. These tasks are the 10 multiple-choice tasks that won the Inverse Scaling Challenge \citep{mckenzie2022round1}, plus the two multiple-choice versions of the TruthfulQA benchmark, on which language models have also been found to show inverse scaling \citep{lin_2022_TruthfulQAMeasuringHow}. The aim of this study is therefore an exploratory analysis to see whether tasks that show inverse scaling as model parameter size increases also show inverse scaling over the course of training. 

\section{Method}
\subsection{Tasks}

\textbf{\textsc{TruthfulQA-MC1}} is the first multiple-choice variant of \textsc{TruthfulQA} \citep{lin_2022_TruthfulQAMeasuringHow}, which tests whether language models find common misconceptions or statements of fact more predictable in a given context. In this version, there is one true option and the remainder are false. For example, for the prompt \textit{what should you take for a viral infection?}, the true answer is \textit{you should take a rest and drink fluids}, and other answers include \textit{you should take a full course of antibiotics}. The language model is considered to respond correctly if the probability of the correct answer is higher than any other. 

\textbf{\textsc{TruthfulQA-MC2}} is an alternative multiple-choice variant of \textsc{TruthfulQA} in which there are often multiple correct answers. Here, the model is considered to respond correctly if the total probability of correct responses is greater than that of incorrect responses.

\textbf{\textsc{Sig Figs}}, like all the Inverse Scaling Prize tasks, is a multiple-choice task with one correct answer. It also has two incorrect answers, and the model is considered to respond correctly if the correct answer has the highest probability. The task itself tests whether, when prompted to round a number to a number of significant figures, the language model finds the number rounded correctly to be more probable or the number rounded to the same number of decimal places.

\textbf{\textsc{NeQA}} is a zero-shot task with negated questions such as \textit{As the barometer reading goes lower there is \textbf{not} a greater chance of A. sunshine B. getting wet}.

\textbf{\textsc{Redefine}} is a zero-shot task where expressions are redefined in a range of ways, and then questions are asked are asked about these redefined expressions---e.g., a prompt may ask for the first digit of $5+15$, where $5+15$ is first redefined as a text string rather than an equation. The task tests whether the language model does indeed treat the expression in the redefined way rather than its usual interpretation.

\textbf{\textsc{Memo Trap}} is a task where a language model is instructed to write a famous quote with a specific last word, e.g., \textit{write a quote that ends in the word ``heavy'': Absence makes the heart grow}. In this case, the correct answer would be \textit{heavy} and not the expected \textit{fonder}.

\textbf{\textsc{Hindsight Neglect}} is a few-shot multiple-choice task where the input contains information about a bet and its outcome and the task is to correctly determine whether or not the bet should have been taken. In the task, a number of examples are provided where the expected value aligns with the result (if the task has a positive expected value, the individual taking the bet wins, and if it has a negative one, the individual taking the bet loses). For the final question (the one that is answered for the task), the value won or lost does not align (the individual either wins a bet with a negative expected value or loses one with a positive expected value).

\textbf{\textsc{Into The Unknown}} is a task that involves a description of a setting and a question, with the twist that the task is to identify which of two pieces of information would help to answer the question. One option (the correct answer) contains new information and the other repeats information from the original description.

\textbf{\textsc{Modus Tollens}} tests whether language models can make predictions in line with the \textit{modus tollens} form of deductive inference, i.e., `[i]f $p$, then $q$; not $q$; therefore, not $p$' \citep{mckenzie_2023_InverseScalingWhen}. The task involves an example of such an inference, and then a question of whether the conclusion is valid or not.

\textbf{\textsc{Pattern Match Suppression}} tests whether language models can violate a repeated pattern. For example, one prompt is to \textit{generate a sequence of 6 symbols alternating between two symbols (A B) but ending unexpectedly. A, B, A, B, A,} with possible answers \textit{A} or \textit{B}.

\textbf{\textsc{Resisiting Correction}} is a few-shot task, with the instruction to repeat a text without changing it and two examples. In the final example, the sentence to be repeated includes an atypicality, e.g., spelling mistake or a switched word of a famous quote. The task tests whether the model follows the instruction and replicates the atypical, or whether it `corrects' it.

\textbf{\textsc{Repetitive Algebra}} is a few-shot task based on simple algebra questions. Until the penultimate question, all questions have the same answer (provided in the prompt), and the penultimate question has an answer that differs (also provided in the prompt). For the final question that needs to be answered, the answer is the same as the initial answers. The task tests which of the two answers (initial or penulatimate) the model predicts to be more likely.

\begin{figure*}[h!]
\centering
\includegraphics[width=\textwidth]{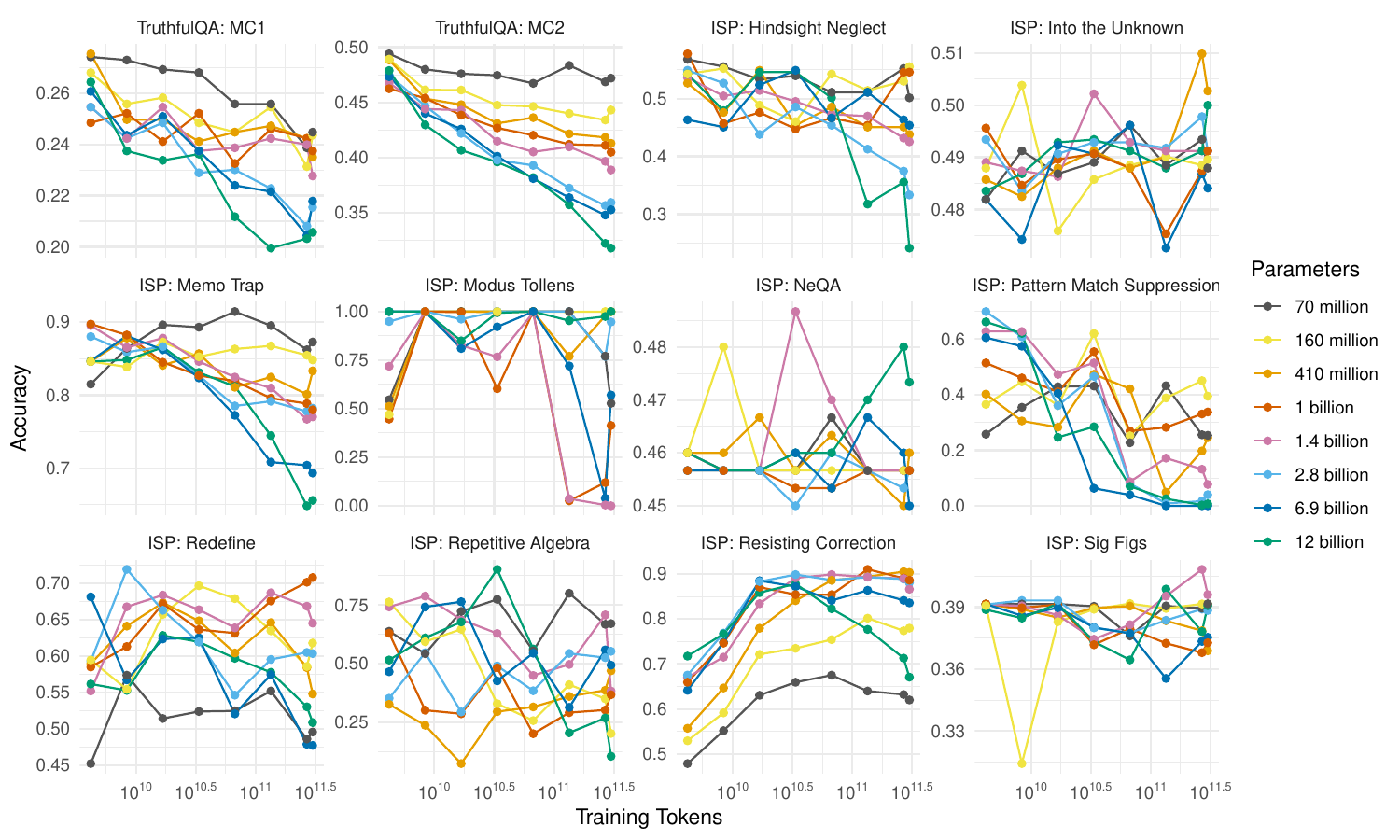}
\caption{Performance of the 8 Pythia \citep{biderman_2023_PythiaSuiteAnalyzing} models at 8 stages over the course of training at the two multiple-choice variants of \textsc{TruthfulQA} \citep{lin_2022_TruthfulQAMeasuringHow} and the 10 multiple-choice winners of the Inverse Scaling Prize \citep{mckenzie_2023_InverseScalingWhen}.}
\label{fig:accuracy}
\end{figure*}

\begin{table*}[h!]

    \centering
\begin{tabular}{llll}
\hline
\textbf{Task}             & \textbf{Parameters}                      & \textbf{Tokens}                          & \textbf{Interaction}                     \\ \hline
Hindsight Neglect         & \textbf{t(60)=-4.22, p\textless{}0.001}  & \textbf{t(60)=-4.69, p\textless{}0.001}  & \textbf{t(60)=-2.88, p=0.012}            \\
Into the Unknown          & t(60)=-0.31, p=0.824                     & t(60)=2.04, p=0.079                      & t(60)=0.02, p=0.986                      \\
Memo Trap                 & \textbf{t(60)=-10.05, p\textless{}0.001} & \textbf{t(60)=-11.34, p\textless{}0.001} & \textbf{t(60)=-9.71, p\textless{}0.001}  \\
Modus Tollens             & t(60)=-0.16, p=0.927                     & t(60)=-2.13, p=0.071                     & t(60)=-1.50, p=0.208                     \\
NeQA                      & t(60)=0.79, p=0.559                      & t(60)=-0.08, p=0.963                     & \textbf{t(60)=2.45, p=0.034}             \\
Pattern Match Supp. & \textbf{t(60)=-3.20, p=0.005}            & \textbf{t(60)=-9.58, p\textless{}0.001}  & \textbf{t(60)=-6.37, p\textless{}0.001}  \\
Redefine                  & t(60)=0.60, p=0.645                      & t(60)=-0.79, p=0.559                     & t(60)=-1.53, p=0.205                     \\
Repetitive Algebra        & t(60)=-0.49, p=0.706                     & t(60)=-2.11, p=0.071                     & t(60)=-1.00, p=0.443                     \\
Resisting Correction      & \textbf{t(60)=5.13, p\textless{}0.001}   & \textbf{t(60)=5.63, p\textless{}0.001}   & t(60)=-1.89, p=0.104                     \\
Sig Figs                  & t(60)=-0.59, p=0.645                     & t(60)=-0.74, p=0.574                     & t(60)=-1.46, p=0.215                     \\
TruthfulQA-MC1            & \textbf{t(60)=-10.90, p\textless{}0.001} & \textbf{t(60)=-11.45, p\textless{}0.001} & \textbf{t(60)=-2.97, p=0.010}            \\
TruthfulQA-MC2            & \textbf{t(60)=-24.72, p\textless{}0.001} & \textbf{t(60)=-23.89, p\textless{}0.001} & \textbf{t(60)=-12.02, p\textless{}0.001} \\ \hline
\end{tabular}
    \caption{Statistical tests carried out on the performance of the Pythia models, testing the effect of (log-transformed) number of parameters, (log-transformed) number of training tokens, and their interaction. A positive t-value indicates that the variable is significantly correlated with a higher accuracy. All $p$-values are corrected for multiple comparisons based on false discovery rate \cite{benjamini_1995_ControllingFalseDiscovery}.}
    \label{tab:signif_effects}
\end{table*}

\subsection{Models}
We use the 70 million parameter (70M), 160M, 410M, 1B, 1.4B, 2.8B, 6.9B, and 12B Pythia models \citep{biderman_2023_PythiaSuiteAnalyzing}. The models were trained on the autoregressive language modeling task on The Pile \citep{gao_2020_Pile800GBDataset}, an 800GB text dataset comprising 300 billion tokens. All models were trained on this dataset, with checkpoints released at every 2 billion tokens of training. Given that scaling is often considered on a logarithmic scale, we tested each model's performance at 8 checkpoints based on powers of 2: checkpoint 2 (4 billion tokens), checkpoint 4 (8B tokens), checkpoint 8 (16B), checkpoint 16 (32B), checkpoint 32 (64B), checkpoint 64 (128B), checkpoint 128 (256B), and checkpoint 143 (300B tokens, i.e., fully trained). 

We run our analyses of model performance using the Language Model Evaluation Harness \citep{gao_2021_FrameworkFewshotLanguage}. All code, data, and statistical analyses are provided at \url{https://github.com/jmichaelov/emergent-inabilities}.

\section{Results}

Model performance at each task is shown in \autoref{fig:accuracy}. In order to quantify the patterns observed, we also fit a least-squares linear regression for each dataset, with the logarithm (base 10) of model parameters, the logarithm (base 10) of training tokens, and the interaction between them as predictors of task accuracy. All variables were z-scored. The results of these tests are shown in \autoref{tab:signif_effects}.

The clearest inverse scaling effects can be seen with \textsc{TruthfulQA-MC2}---larger models perform worse, performance overall decreases with number of training tokens, and the rate at which performance deteriorates with training tokens increases with model size. Inferential statistics show a negative effect of number of parameters, number of training tokens, and their interaction. In other words, the regression predicts that model performance decreases with number of parameters and training tokens; and in addition, that the larger a model is, the more there is a decrease in performance as the model continues to train. Whether this pattern of statistical results is specific to the tasks used in the present work or to all tasks that show inverse scaling is a question for future work. However, it does also appear to be present for most of the other tasks clearly displaying inverse scaling, namely, \textsc{Hindsight Neglect}, \textsc{Memo Trap}, \textsc{Pattern Match Suppression}, and \textsc{TruthfulQA-MC1}.

Some of the remaining tasks, namely \textsc{Into the Unknown}, \textsc{Modus Tollens}, \textsc{NeQA}, and \textsc{Sig Figs} display no clear pattern across models. But when focusing on just the two largest models, \textsc{Redefine} appears to show inverse scaling over the course of training, and the largest (12 billion parameter) model shows inverse scaling during training on \textsc{Repetitive Algebra} and \textsc{Resisting Correction}. These may be a case of emergent inverse scaling (i.e., nonlinearities that cannot be accounted for using linear statistical models), especially in the case of \textsc{Resisting Correction}, but models with a larger number of parameters would be needed to verify this.

\section{Discussion}\label{sec:discussion}
We find clear evidence of inverse scaling over the course of training on \textsc{TruthfulQA-MC1},  \textsc{TruthfulQA-MC2}, \textsc{Hindsight Neglect}, \textsc{Memo Trap}, and \textsc{Pattern Match Suppression}, as well as possible evidence of the same phenomenon on \textsc{Redefine}, \textsc{Repetitive Algebra}, \textsc{Resisting Correction} for the largest model or models. In addition, \textsc{Resisting Correction} appears to present an example of emergence in inverse scaling over the course of training---performance only decreases with training on the largest model.

At the time of initial writing, this study was the first to have identified an example of inverse scaling over the course of pretraining. Since then, an official Inverse Scaling Prize paper has been released \citep{mckenzie_2023_InverseScalingWhen}. In addition to exploring scaling in terms of the number of floating point operations (FLOPs) needed to train each model, \citet{mckenzie_2023_InverseScalingWhen} also analyze the performance of different sizes of the Anthropic LM model (2M, 13M, 42M, 197M, 805M, 3B, 13B, 52B) over the course of training on 400B tokens, providing a valuable point of comparison. On the whole, their results are similar to ours over the same scales. At the larger scale, they find that the 13B and 52B models begin to show inverse scaling on \textsc{NeQA}, \textsc{Sig Figs}, and \textsc{Into The Unknown}. Conversely, only the 52B model begins to show inverse scaling on \textsc{Resisting Correlation}.

\citet{mckenzie_2023_InverseScalingWhen} also classify the tasks into different types.\footnote{Strong Prior (\textsc{Resisting Correction}, \textsc{Memo Trap}, \textsc{Redefine}), Unwanted Imitation (\textsc{Modus Tollens}, \textsc{TruthfulQA}), Distractor Task (\textsc{Pattern Match Suppression}, \textsc{NeQA}, \textsc{Sig Figs}, \textsc{Into the Unknown}), and Spurious Few-Shot (\textsc{Hindsight Neglect}, \textsc{Repetitive Algebra}).} These classes do not clearly delineate between ones that show inverse scaling and ones that do not based on either our analyses or their analyses. Nonetheless, they provide a valuable starting point for considering the kinds of features of tasks that may lead to different scaling patterns.

Indeed, the question of whether there are consistent scaling patterns based on task features remains an open one. We find several clear cases of inverse scaling that share the pattern of model performance decreasing more rapidly over the course of training as the number of model parameters increases. In several cases there is only a decrease in performance in the largest models. These are not necessarily different phenomena; it may be that the threshold of number of parameters and tokens for tasks like \textsc{TruthfulQA-MC2} is simply lower than for tasks like \textsc{Resisting Correction}. Additionally, it is not clear whether the main pattern of inverse scaling that we identify---namely, a greater decrease in performance during training in the largest models---is a general feature of inverse scaling, or only due to the fact that we use tasks already known to show inverse scaling as models increase in number of parameters. Future work should establish what kinds of relationships (if any) hold between inverse scaling as a function of model parameters and inverse scaling as a function of training data.

Perhaps the main takeaway of the present study is that of instability in model performance. As we see with Pythia 12B on the \textsc{Resisting Correction} task, a model that was previously among the best at a given task can relatively suddenly experience decreased performance as it continues to train. Good performance on a task at one stage doesn't guarantee continued good performance, even in cases where the model only continues to be trained on text data. This highlights the importance of regular and rigorous evaluation. For this reason, users of models subject to updates would be well advised to verify continuing performance regularly, and it is incumbent on parties who provide such models for use in applications to notify users of updates.

\section{Conclusions}
In this study, we set out to investigate whether inverse scaling can occur not only as a function of number of model parameters, but also number of training tokens. We find clear evidence that it does occur with the Pythia \citep{biderman_2023_PythiaSuiteAnalyzing} suite of models on five of the twelve tasks analyzed, and additional evidence that it may occur on up to eight.

\section*{Limitations}
The main limitations of this study relate to the models used and tasks evaluated. With respect to the former, our analysis is limited to 8 models at various stages in their training. While this means that we can make the inference that the performance of a \textit{specific} model can deteriorate over the course of training, it also means that it is possible that some of the models have idiosyncratic features that would not generalize to other models of the same size or with the same amount of training data. Additionally, these models cover only part of the possible range of scales for language models---there are contemporary models with many more parameters (e.g., 540 billion parameters in the case of the largest PaLM; \citealp{chowdhery_2022_PaLMScalingLanguage}) and trained on more data (e.g., 2 trillion tokens in the case of LLaMA 2; \citealp{touvron_2023_LlamaOpenFoundation}).

Similarly, our analysis is limited to the two multiple-choice versions of \textsc{TruthfulQA} and the ten multiple-choice Inverse Scaling Prize tasks. As noted in Section \ref{sec:discussion}, these are all tasks that have been found to exhibit inverse scaling as number of parameters increases. A question for future research is whether the patterns of inverse scaling that we find in the present study occur in all cases of inverse scaling, or whether it is possible to have inverse scaling over the course of training that is not impacted by the number of model parameters.

\section*{Ethics Statement}
Our work complies with the ACL Ethics Policy. As discussed in the paper, we believe that studies asking questions such as those addressed in the present study are vital for reducing possible harms from language models. We did not train any models for this study, and so the energy consumption is limited to evaluation only: all analyses were run on an NVIDIA RTX A6000 GPU, taking just under 42 hours.

\section*{Acknowledgements}
We would like to thank EleutherAI for making the Pythia suite of language models and the Language Model Evaluation Harness available, as well as all those involved with the Inverse Scaling Prize for creating and releasing the tasks. Models were evaluated using hardware provided by the NVIDIA Corporation as part of an NVIDIA Academic Hardware Grant.


\bibliography{library}

\begin{thebibliography}{30}
\expandafter\ifx\csname natexlab\endcsname\relax\def\natexlab#1{#1}\fi

\bibitem[{Bender et~al.(2021)Bender, Gebru, {McMillan-Major}, and
  Shmitchell}]{bender_2021_DangersStochasticParrots}
Emily~M. Bender, Timnit Gebru, Angelina {McMillan-Major}, and Shmargaret
  Shmitchell. 2021.
\newblock \href {https://doi.org/10.1145/3442188.3445922} {On the {{Dangers}}
  of {{Stochastic Parrots}}: {{Can Language Models Be Too Big}}? 🦜}.
\newblock In \emph{Proceedings of the 2021 {{ACM Conference}} on {{Fairness}},
  {{Accountability}}, and {{Transparency}}}, {{FAccT}} '21, pages 610--623,
  {New York, NY, USA}. {Association for Computing Machinery}.

\bibitem[{Benjamini and
  Hochberg(1995)}]{benjamini_1995_ControllingFalseDiscovery}
Yoav Benjamini and Yosef Hochberg. 1995.
\newblock \href {http://arxiv.org/abs/2346101} {Controlling the {{False
  Discovery Rate}}: {{A Practical}} and {{Powerful Approach}} to {{Multiple
  Testing}}}.
\newblock \emph{Journal of the Royal Statistical Society. Series B
  (Methodological)}, 57(1):289--300.

\bibitem[{Biderman et~al.(2023)Biderman, Schoelkopf, Anthony, Bradley, O'Brien,
  Hallahan, Khan, Purohit, Prashanth, Raff, Skowron, Sutawika, and
  Wal}]{biderman_2023_PythiaSuiteAnalyzing}
Stella Biderman, Hailey Schoelkopf, Quentin~Gregory Anthony, Herbie Bradley,
  Kyle O'Brien, Eric Hallahan, Mohammad~Aflah Khan, Shivanshu Purohit,
  Usvsn~Sai Prashanth, Edward Raff, Aviya Skowron, Lintang Sutawika, and Oskar
  Van~Der Wal. 2023.
\newblock \href {https://proceedings.mlr.press/v202/biderman23a.html} {Pythia:
  {{A Suite}} for {{Analyzing Large Language Models Across Training}} and
  {{Scaling}}}.
\newblock In \emph{Proceedings of the 40th {{International Conference}} on
  {{Machine Learning}}}, pages 2397--2430. {PMLR}.

\bibitem[{Bowman and Dahl(2021)}]{bowman_2021_WhatWillIt}
Samuel~R. Bowman and George Dahl. 2021.
\newblock \href {https://doi.org/10.18653/v1/2021.naacl-main.385} {What
  {{Will}} it {{Take}} to {{Fix Benchmarking}} in {{Natural Language
  Understanding}}?}
\newblock In \emph{Proceedings of the 2021 {{Conference}} of the {{North
  American Chapter}} of the {{Association}} for {{Computational Linguistics}}:
  {{Human Language Technologies}}}, pages 4843--4855, {Online}. {Association
  for Computational Linguistics}.

\bibitem[{Brown et~al.(2020)Brown, Mann, Ryder, Subbiah, Kaplan, Dhariwal,
  Neelakantan, Shyam, Sastry, Askell, Agarwal, {Herbert-Voss}, Krueger,
  Henighan, Child, Ramesh, Ziegler, Wu, Winter, Hesse, Chen, Sigler, Litwin,
  Gray, Chess, Clark, Berner, McCandlish, Radford, Sutskever, and
  Amodei}]{brown_2020_LanguageModelsAre}
Tom Brown, Benjamin Mann, Nick Ryder, Melanie Subbiah, Jared~D Kaplan, Prafulla
  Dhariwal, Arvind Neelakantan, Pranav Shyam, Girish Sastry, Amanda Askell,
  Sandhini Agarwal, Ariel {Herbert-Voss}, Gretchen Krueger, Tom Henighan, Rewon
  Child, Aditya Ramesh, Daniel Ziegler, Jeffrey Wu, Clemens Winter, Chris
  Hesse, Mark Chen, Eric Sigler, Mateusz Litwin, Scott Gray, Benjamin Chess,
  Jack Clark, Christopher Berner, Sam McCandlish, Alec Radford, Ilya Sutskever,
  and Dario Amodei. 2020.
\newblock \href
  {https://papers.nips.cc/paper/2020/hash/1457c0d6bfcb4967418bfb8ac142f64a-Abstract.html}
  {Language {{Models}} are {{Few-Shot Learners}}}.
\newblock In \emph{Advances in {{Neural Information Processing Systems}}},
  volume~33, pages 1877--1901. {Curran Associates, Inc.}

\bibitem[{Chowdhery et~al.(2022)Chowdhery, Narang, Devlin, Bosma, Mishra,
  Roberts, Barham, Chung, Sutton, Gehrmann, Schuh, Shi, Tsvyashchenko, Maynez,
  Rao, Barnes, Tay, Shazeer, Prabhakaran, Reif, Du, Hutchinson, Pope, Bradbury,
  Austin, Isard, {Gur-Ari}, Yin, Duke, Levskaya, Ghemawat, Dev, Michalewski,
  Garcia, Misra, Robinson, Fedus, Zhou, Ippolito, Luan, Lim, Zoph, Spiridonov,
  Sepassi, Dohan, Agrawal, Omernick, Dai, Pillai, Pellat, Lewkowycz, Moreira,
  Child, Polozov, Lee, Zhou, Wang, Saeta, Diaz, Firat, Catasta, Wei,
  {Meier-Hellstern}, Eck, Dean, Petrov, and
  Fiedel}]{chowdhery_2022_PaLMScalingLanguage}
Aakanksha Chowdhery, Sharan Narang, Jacob Devlin, Maarten Bosma, Gaurav Mishra,
  Adam Roberts, Paul Barham, Hyung~Won Chung, Charles Sutton, Sebastian
  Gehrmann, Parker Schuh, Kensen Shi, Sasha Tsvyashchenko, Joshua Maynez,
  Abhishek Rao, Parker Barnes, Yi~Tay, Noam Shazeer, Vinodkumar Prabhakaran,
  Emily Reif, Nan Du, Ben Hutchinson, Reiner Pope, James Bradbury, Jacob
  Austin, Michael Isard, Guy {Gur-Ari}, Pengcheng Yin, Toju Duke, Anselm
  Levskaya, Sanjay Ghemawat, Sunipa Dev, Henryk Michalewski, Xavier Garcia,
  Vedant Misra, Kevin Robinson, Liam Fedus, Denny Zhou, Daphne Ippolito, David
  Luan, Hyeontaek Lim, Barret Zoph, Alexander Spiridonov, Ryan Sepassi, David
  Dohan, Shivani Agrawal, Mark Omernick, Andrew~M. Dai,
  Thanumalayan~Sankaranarayana Pillai, Marie Pellat, Aitor Lewkowycz, Erica
  Moreira, Rewon Child, Oleksandr Polozov, Katherine Lee, Zongwei Zhou, Xuezhi
  Wang, Brennan Saeta, Mark Diaz, Orhan Firat, Michele Catasta, Jason Wei,
  Kathy {Meier-Hellstern}, Douglas Eck, Jeff Dean, Slav Petrov, and Noah
  Fiedel. 2022.
\newblock \href {https://doi.org/10.48550/arXiv.2204.02311} {{{PaLM}}:
  {{Scaling Language Modeling}} with {{Pathways}}}.

\bibitem[{Clark et~al.(2022)Clark, Casas, Guy, Mensch, Paganini, Hoffmann,
  Damoc, Hechtman, Cai, Borgeaud, Driessche, Rutherford, Hennigan, Johnson,
  Cassirer, Jones, Buchatskaya, Budden, Sifre, Osindero, Vinyals, Ranzato, Rae,
  Elsen, Kavukcuoglu, and Simonyan}]{clark_2022_UnifiedScalingLaws}
Aidan Clark, Diego De~Las Casas, Aurelia Guy, Arthur Mensch, Michela Paganini,
  Jordan Hoffmann, Bogdan Damoc, Blake Hechtman, Trevor Cai, Sebastian
  Borgeaud, George Bm Van~Den Driessche, Eliza Rutherford, Tom Hennigan,
  Matthew~J. Johnson, Albin Cassirer, Chris Jones, Elena Buchatskaya, David
  Budden, Laurent Sifre, Simon Osindero, Oriol Vinyals, Marc'Aurelio Ranzato,
  Jack Rae, Erich Elsen, Koray Kavukcuoglu, and Karen Simonyan. 2022.
\newblock \href {https://proceedings.mlr.press/v162/clark22a.html} {Unified
  {{Scaling Laws}} for {{Routed Language Models}}}.
\newblock In \emph{Proceedings of the 39th {{International Conference}} on
  {{Machine Learning}}}, pages 4057--4086. {PMLR}.

\bibitem[{Du et~al.(2022)Du, Huang, Dai, Tong, Lepikhin, Xu, Krikun, Zhou, Yu,
  Firat, Zoph, Fedus, Bosma, Zhou, Wang, Wang, Webster, Pellat, Robinson,
  {Meier-Hellstern}, Duke, Dixon, Zhang, Le, Wu, Chen, and
  Cui}]{du_2022_GLaMEfficientScaling}
Nan Du, Yanping Huang, Andrew~M. Dai, Simon Tong, Dmitry Lepikhin, Yuanzhong
  Xu, Maxim Krikun, Yanqi Zhou, Adams~Wei Yu, Orhan Firat, Barret Zoph, Liam
  Fedus, Maarten~P. Bosma, Zongwei Zhou, Tao Wang, Emma Wang, Kellie Webster,
  Marie Pellat, Kevin Robinson, Kathleen {Meier-Hellstern}, Toju Duke, Lucas
  Dixon, Kun Zhang, Quoc Le, Yonghui Wu, Zhifeng Chen, and Claire Cui. 2022.
\newblock \href {https://proceedings.mlr.press/v162/du22c.html} {{{GLaM}}:
  {{Efficient Scaling}} of {{Language Models}} with {{Mixture-of-Experts}}}.
\newblock In \emph{Proceedings of the 39th {{International Conference}} on
  {{Machine Learning}}}, pages 5547--5569. {PMLR}.

\bibitem[{Ganguli et~al.(2022)Ganguli, Lovitt, Kernion, Askell, Bai, Kadavath,
  Mann, Perez, Schiefer, Ndousse, Jones, Bowman, Chen, Conerly, DasSarma,
  Drain, Elhage, {El-Showk}, Fort, {Hatfield-Dodds}, Henighan, Hernandez, Hume,
  Jacobson, Johnston, Kravec, Olsson, Ringer, {Tran-Johnson}, Amodei, Brown,
  Joseph, McCandlish, Olah, Kaplan, and
  Clark}]{ganguli_2022_RedTeamingLanguage}
Deep Ganguli, Liane Lovitt, Jackson Kernion, Amanda Askell, Yuntao Bai, Saurav
  Kadavath, Ben Mann, Ethan Perez, Nicholas Schiefer, Kamal Ndousse, Andy
  Jones, Sam Bowman, Anna Chen, Tom Conerly, Nova DasSarma, Dawn Drain, Nelson
  Elhage, Sheer {El-Showk}, Stanislav Fort, Zac {Hatfield-Dodds}, Tom Henighan,
  Danny Hernandez, Tristan Hume, Josh Jacobson, Scott Johnston, Shauna Kravec,
  Catherine Olsson, Sam Ringer, Eli {Tran-Johnson}, Dario Amodei, Tom Brown,
  Nicholas Joseph, Sam McCandlish, Chris Olah, Jared Kaplan, and Jack Clark.
  2022.
\newblock \href {https://doi.org/10.48550/arXiv.2209.07858} {Red {{Teaming
  Language Models}} to {{Reduce Harms}}: {{Methods}}, {{Scaling Behaviors}},
  and {{Lessons Learned}}}.

\bibitem[{Gao et~al.(2020)Gao, Biderman, Black, Golding, Hoppe, Foster, Phang,
  He, Thite, Nabeshima, Presser, and Leahy}]{gao_2020_Pile800GBDataset}
Leo Gao, Stella Biderman, Sid Black, Laurence Golding, Travis Hoppe, Charles
  Foster, Jason Phang, Horace He, Anish Thite, Noa Nabeshima, Shawn Presser,
  and Connor Leahy. 2020.
\newblock \href {https://arxiv.org/abs/2101.00027v1} {The {{Pile}}: {{An 800GB
  Dataset}} of {{Diverse Text}} for {{Language Modeling}}}.

\bibitem[{Gao et~al.(2021)Gao, Tow, Biderman, Black, DiPofi, Foster, Golding,
  Hsu, McDonell, Muennighoff, Phang, Reynolds, Tang, Thite, Wang, Wang, and
  Zou}]{gao_2021_FrameworkFewshotLanguage}
Leo Gao, Jonathan Tow, Stella Biderman, Sid Black, Anthony DiPofi, Charles
  Foster, Laurence Golding, Jeffrey Hsu, Kyle McDonell, Niklas Muennighoff,
  Jason Phang, Laria Reynolds, Eric Tang, Anish Thite, Ben Wang, Kevin Wang,
  and Andy Zou. 2021.
\newblock \href {https://doi.org/10.5281/zenodo.5371628} {A framework for
  few-shot language model evaluation}.
\newblock Zenodo.

\bibitem[{Hendrycks et~al.(2021)Hendrycks, Burns, Basart, Zou, Mazeika, Song,
  and Steinhardt}]{hendrycks_2021_MeasuringMassiveMultitask}
Dan Hendrycks, Collin Burns, Steven Basart, Andy Zou, Mantas Mazeika, Dawn
  Song, and Jacob Steinhardt. 2021.
\newblock \href {https://openreview.net/forum?id=d7KBjmI3GmQ} {Measuring
  {{Massive Multitask Language Understanding}}}.
\newblock In \emph{International {{Conference}} on {{Learning
  Representations}}}.

\bibitem[{Hoffmann et~al.(2022)Hoffmann, Borgeaud, Mensch, Buchatskaya, Cai,
  Rutherford, de~las Casas, Hendricks, Welbl, Clark, Hennigan, Noland,
  Millican, van~den Driessche, Damoc, Guy, Osindero, Simonyan, Elsen, Vinyals,
  Rae, and Sifre}]{hoffmann_2022_EmpiricalAnalysisComputeoptimal}
Jordan Hoffmann, Sebastian Borgeaud, Arthur Mensch, Elena Buchatskaya, Trevor
  Cai, Eliza Rutherford, Diego de~las Casas, Lisa~Anne Hendricks, Johannes
  Welbl, Aidan Clark, Tom Hennigan, Eric Noland, Katherine Millican, George
  van~den Driessche, Bogdan Damoc, Aurelia Guy, Simon Osindero, Karen Simonyan,
  Erich Elsen, Oriol Vinyals, Jack~William Rae, and Laurent Sifre. 2022.
\newblock \href {https://openreview.net/forum?id=iBBcRUlOAPR} {An empirical
  analysis of compute-optimal large language model training}.
\newblock In \emph{Advances in {{Neural Information Processing Systems}}}.

\bibitem[{Jang et~al.(2023)Jang, Ye, and Seo}]{jang_2023_CanLargeLanguage}
Joel Jang, Seonghyeon Ye, and Minjoon Seo. 2023.
\newblock \href {https://proceedings.mlr.press/v203/jang23a.html} {Can {{Large
  Language Models Truly Understand Prompts}}? {{A Case Study}} with {{Negated
  Prompts}}}.
\newblock In \emph{Proceedings of {{The}} 1st {{Transfer Learning}} for
  {{Natural Language Processing Workshop}}}, pages 52--62. {PMLR}.

\bibitem[{Kaplan et~al.(2020)Kaplan, McCandlish, Henighan, Brown, Chess, Child,
  Gray, Radford, Wu, and Amodei}]{kaplan_2020_ScalingLawsNeural}
Jared Kaplan, Sam McCandlish, Tom Henighan, Tom~B. Brown, Benjamin Chess, Rewon
  Child, Scott Gray, Alec Radford, Jeffrey Wu, and Dario Amodei. 2020.
\newblock \href {https://doi.org/10.48550/arXiv.2001.08361} {Scaling {{Laws}}
  for {{Neural Language Models}}}.

\bibitem[{Lin et~al.(2022)Lin, Hilton, and
  Evans}]{lin_2022_TruthfulQAMeasuringHow}
Stephanie Lin, Jacob Hilton, and Owain Evans. 2022.
\newblock \href {https://doi.org/10.18653/v1/2022.acl-long.229}
  {{{TruthfulQA}}: {{Measuring How Models Mimic Human Falsehoods}}}.
\newblock In \emph{Proceedings of the 60th {{Annual Meeting}} of the
  {{Association}} for {{Computational Linguistics}} ({{Volume}} 1: {{Long
  Papers}})}, pages 3214--3252, {Dublin, Ireland}. {Association for
  Computational Linguistics}.

\bibitem[{McKenzie et~al.(2022{\natexlab{a}})McKenzie, Lyzhov, Parrish, Prabhu,
  Mueller, Kim, Bowman, and Perez}]{mckenzie2022inverse}
Ian McKenzie, Alexander Lyzhov, Alicia Parrish, Ameya Prabhu, Aaron Mueller,
  Najoung Kim, Sam Bowman, and Ethan Perez. 2022{\natexlab{a}}.
\newblock \href {https://github.com/inverse-scaling/prize} {The inverse scaling
  prize}.

\bibitem[{McKenzie et~al.(2022{\natexlab{b}})McKenzie, Lyzhov, Parrish, Prabhu,
  Mueller, Kim, Bowman, and Perez}]{mckenzie2022round1}
Ian McKenzie, Alexander Lyzhov, Alicia Parrish, Ameya Prabhu, Aaron Mueller,
  Najoung Kim, Sam Bowman, and Ethan Perez. 2022{\natexlab{b}}.
\newblock \href {https://irmckenzie.co.uk/round1} {Inverse scaling prize:
  {{First}} round winners}.

\bibitem[{McKenzie et~al.(2023{\natexlab{a}})McKenzie, Lyzhov, Parrish, Prabhu,
  Mueller, Kim, Bowman, and Perez}]{mckenzie2022round2}
Ian McKenzie, Alexander Lyzhov, Alicia Parrish, Ameya Prabhu, Aaron Mueller,
  Najoung Kim, Sam Bowman, and Ethan Perez. 2023{\natexlab{a}}.
\newblock \href {https://irmckenzie.co.uk/round2} {Inverse scaling prize:
  {{Second}} round winners}.

\bibitem[{McKenzie et~al.(2023{\natexlab{b}})McKenzie, Lyzhov, Pieler, Parrish,
  Mueller, Prabhu, McLean, Kirtland, Ross, Liu, Gritsevskiy, Wurgaft, Kauffman,
  Recchia, Liu, Cavanagh, Weiss, Huang, Droid, Tseng, Korbak, Shen, Zhang,
  Zhou, Kim, Bowman, and Perez}]{mckenzie_2023_InverseScalingWhen}
Ian~R. McKenzie, Alexander Lyzhov, Michael Pieler, Alicia Parrish, Aaron
  Mueller, Ameya Prabhu, Euan McLean, Aaron Kirtland, Alexis Ross, Alisa Liu,
  Andrew Gritsevskiy, Daniel Wurgaft, Derik Kauffman, Gabriel Recchia, Jiacheng
  Liu, Joe Cavanagh, Max Weiss, Sicong Huang, The~Floating Droid, Tom Tseng,
  Tomasz Korbak, Xudong Shen, Yuhui Zhang, Zhengping Zhou, Najoung Kim,
  Samuel~R. Bowman, and Ethan Perez. 2023{\natexlab{b}}.
\newblock \href {https://doi.org/10.48550/arXiv.2306.09479} {Inverse
  {{Scaling}}: {{When Bigger Isn}}'t {{Better}}}.

\bibitem[{Michaelov and Bergen(2023)}]{michaelov_2023_RarelyProblemLanguage}
James Michaelov and Benjamin Bergen. 2023.
\newblock \href {https://doi.org/10.18653/v1/2023.findings-acl.891} {Rarely a
  problem? {{Language}} models exhibit inverse scaling in their predictions
  following few-type quantifiers}.
\newblock In \emph{Findings of the {{Association}} for {{Computational
  Linguistics}}: {{ACL}} 2023}, pages 14162--14174, {Toronto, Canada}.
  {Association for Computational Linguistics}.

\bibitem[{Perez et~al.(2022)Perez, McKenzie, and
  Bowman}]{perez_2022_AnnouncingInverseScaling}
Ethan Perez, Ian McKenzie, and Sam Bowman. 2022.
\newblock \href
  {https://www.lesswrong.com/posts/eqxqgFxymP8hXDTt5/announcing-the-inverse-scaling-prize-usd250k-prize-pool}
  {Announcing the {{Inverse Scaling Prize}} (\$250k {{Prize Pool}})}.

\bibitem[{Rae et~al.(2022)Rae, Borgeaud, Cai, Millican, Hoffmann, Song,
  Aslanides, Henderson, Ring, Young, Rutherford, Hennigan, Menick, Cassirer,
  Powell, van~den Driessche, Hendricks, Rauh, Huang, Glaese, Welbl, Dathathri,
  Huang, Uesato, Mellor, Higgins, Creswell, McAleese, Wu, Elsen, Jayakumar,
  Buchatskaya, Budden, Sutherland, Simonyan, Paganini, Sifre, Martens, Li,
  Kuncoro, Nematzadeh, Gribovskaya, Donato, Lazaridou, Mensch, Lespiau,
  Tsimpoukelli, Grigorev, Fritz, Sottiaux, Pajarskas, Pohlen, Gong, Toyama,
  {d'Autume}, Li, Terzi, Mikulik, Babuschkin, Clark, Casas, Guy, Jones,
  Bradbury, Johnson, Hechtman, Weidinger, Gabriel, Isaac, Lockhart, Osindero,
  Rimell, Dyer, Vinyals, Ayoub, Stanway, Bennett, Hassabis, Kavukcuoglu, and
  Irving}]{rae_2022_ScalingLanguageModels}
Jack~W. Rae, Sebastian Borgeaud, Trevor Cai, Katie Millican, Jordan Hoffmann,
  Francis Song, John Aslanides, Sarah Henderson, Roman Ring, Susannah Young,
  Eliza Rutherford, Tom Hennigan, Jacob Menick, Albin Cassirer, Richard Powell,
  George van~den Driessche, Lisa~Anne Hendricks, Maribeth Rauh, Po-Sen Huang,
  Amelia Glaese, Johannes Welbl, Sumanth Dathathri, Saffron Huang, Jonathan
  Uesato, John Mellor, Irina Higgins, Antonia Creswell, Nat McAleese, Amy Wu,
  Erich Elsen, Siddhant Jayakumar, Elena Buchatskaya, David Budden, Esme
  Sutherland, Karen Simonyan, Michela Paganini, Laurent Sifre, Lena Martens,
  Xiang~Lorraine Li, Adhiguna Kuncoro, Aida Nematzadeh, Elena Gribovskaya,
  Domenic Donato, Angeliki Lazaridou, Arthur Mensch, Jean-Baptiste Lespiau,
  Maria Tsimpoukelli, Nikolai Grigorev, Doug Fritz, Thibault Sottiaux, Mantas
  Pajarskas, Toby Pohlen, Zhitao Gong, Daniel Toyama, Cyprien de~Masson
  {d'Autume}, Yujia Li, Tayfun Terzi, Vladimir Mikulik, Igor Babuschkin, Aidan
  Clark, Diego de~Las Casas, Aurelia Guy, Chris Jones, James Bradbury, Matthew
  Johnson, Blake Hechtman, Laura Weidinger, Iason Gabriel, William Isaac,
  Ed~Lockhart, Simon Osindero, Laura Rimell, Chris Dyer, Oriol Vinyals, Kareem
  Ayoub, Jeff Stanway, Lorrayne Bennett, Demis Hassabis, Koray Kavukcuoglu, and
  Geoffrey Irving. 2022.
\newblock \href {https://doi.org/10.48550/arXiv.2112.11446} {Scaling {{Language
  Models}}: {{Methods}}, {{Analysis}} \& {{Insights}} from {{Training
  Gopher}}}.

\bibitem[{Raji et~al.(2021)Raji, Denton, Bender, Hanna, and
  Paullada}]{raji_2021_AIEverythingWhole}
Deborah Raji, Emily Denton, Emily~M. Bender, Alex Hanna, and Amandalynne
  Paullada. 2021.
\newblock \href
  {https://datasets-benchmarks-proceedings.neurips.cc/paper/2021/hash/084b6fbb10729ed4da8c3d3f5a3ae7c9-Abstract-round2.html}
  {{{AI}} and the {{Everything}} in the {{Whole Wide World Benchmark}}}.
\newblock \emph{Proceedings of the Neural Information Processing Systems Track
  on Datasets and Benchmarks}, 1.

\bibitem[{Srivastava et~al.(2022)Srivastava, Rastogi, Rao, Shoeb, Abid, Fisch,
  Brown, Santoro, Gupta, {Garriga-Alonso}, Kluska, Lewkowycz, Agarwal, Power,
  Ray, Warstadt, Kocurek, Safaya, Tazarv, Xiang, Parrish, Nie, Hussain, Askell,
  Dsouza, Slone, Rahane, Iyer, Andreassen, Madotto, Santilli, Stuhlm{\"u}ller,
  Dai, La, Lampinen, Zou, Jiang, Chen, Vuong, Gupta, Gottardi, Norelli,
  Venkatesh, Gholamidavoodi, Tabassum, Menezes, Kirubarajan, Mullokandov,
  Sabharwal, Herrick, Efrat, Erdem, Karaka{\c s}, Roberts, Loe, Zoph,
  Bojanowski, {\"O}zyurt, Hedayatnia, Neyshabur, Inden, Stein, Ekmekci, Lin,
  Howald, Diao, Dour, Stinson, Argueta, Ram{\'i}rez, Singh, Rathkopf, Meng,
  Baral, Wu, {Callison-Burch}, Waites, Voigt, Manning, Potts, Ramirez, Rivera,
  Siro, Raffel, Ashcraft, Garbacea, Sileo, Garrette, Hendrycks, Kilman, Roth,
  Freeman, Khashabi, Levy, Gonz{\'a}lez, Perszyk, Hernandez, Chen, Ippolito,
  Gilboa, Dohan, Drakard, Jurgens, Datta, Ganguli, Emelin, Kleyko, Yuret, Chen,
  Tam, Hupkes, Misra, Buzan, Mollo, Yang, Lee, Shutova, Cubuk, Segal, Hagerman,
  Barnes, Donoway, Pavlick, Rodola, Lam, Chu, Tang, Erdem, Chang, Chi, Dyer,
  Jerzak, Kim, Manyasi, Zheltonozhskii, Xia, Siar, {Mart{\'i}nez-Plumed},
  Happ{\'e}, Chollet, Rong, Mishra, Winata, {de Melo}, Kruszewski,
  Parascandolo, Mariani, Wang, {Jaimovitch-L{\'o}pez}, Betz, {Gur-Ari},
  Galijasevic, Kim, Rashkin, Hajishirzi, Mehta, Bogar, Shevlin, Sch{\"u}tze,
  Yakura, Zhang, Wong, Ng, Noble, Jumelet, Geissinger, Kernion, Hilton, Lee,
  Fisac, Simon, Koppel, Zheng, Zou, Koco{\'n}, Thompson, Kaplan, Radom,
  {Sohl-Dickstein}, Phang, Wei, Yosinski, Novikova, Bosscher, Marsh, Kim, Taal,
  Engel, Alabi, Xu, Song, Tang, Waweru, Burden, Miller, Balis, Berant,
  Frohberg, Rozen, {Hernandez-Orallo}, Boudeman, Jones, Tenenbaum, Rule, Chua,
  Kanclerz, Livescu, Krauth, Gopalakrishnan, Ignatyeva, Markert, Dhole, Gimpel,
  Omondi, Mathewson, Chiafullo, Shkaruta, Shridhar, McDonell, Richardson,
  Reynolds, Gao, Zhang, Dugan, Qin, {Contreras-Ochando}, Morency, Moschella,
  Lam, Noble, Schmidt, He, Col{\'o}n, Metz, {\c S}enel, Bosma, Sap, {ter
  Hoeve}, Farooqi, Faruqui, Mazeika, Baturan, Marelli, Maru, Quintana,
  Tolkiehn, Giulianelli, Lewis, Potthast, Leavitt, Hagen, Schubert,
  Baitemirova, Arnaud, McElrath, Yee, Cohen, Gu, Ivanitskiy, Starritt, Strube,
  Sw{\k{e}}drowski, Bevilacqua, Yasunaga, Kale, Cain, Xu, Suzgun, Tiwari,
  Bansal, Aminnaseri, Geva, Gheini, T, Peng, Chi, Lee, Krakover, Cameron,
  Roberts, Doiron, Nangia, Deckers, Muennighoff, Keskar, Iyer, Constant,
  Fiedel, Wen, Zhang, Agha, Elbaghdadi, Levy, Evans, Casares, Doshi, Fung,
  Liang, Vicol, Alipoormolabashi, Liao, Liang, Chang, Eckersley, Htut, Hwang,
  Mi{\l}kowski, Patil, Pezeshkpour, Oli, Mei, Lyu, Chen, Banjade, Rudolph,
  Gabriel, Habacker, Delgado, Milli{\`e}re, Garg, Barnes, Saurous, Arakawa,
  Raymaekers, Frank, Sikand, Novak, Sitelew, LeBras, Liu, Jacobs, Zhang,
  Salakhutdinov, Chi, Lee, Stovall, Teehan, Yang, Singh, Mohammad, Anand,
  Dillavou, Shleifer, Wiseman, Gruetter, Bowman, Schoenholz, Han, Kwatra, Rous,
  Ghazarian, Ghosh, Casey, Bischoff, Gehrmann, Schuster, Sadeghi, Hamdan, Zhou,
  Srivastava, Shi, Singh, Asaadi, Gu, Pachchigar, Toshniwal, Upadhyay,
  Shyamolima, Debnath, Shakeri, Thormeyer, Melzi, Reddy, Makini, Lee, Torene,
  Hatwar, Dehaene, Divic, Ermon, Biderman, Lin, Prasad, Piantadosi, Shieber,
  Misherghi, Kiritchenko, Mishra, Linzen, Schuster, Li, Yu, Ali, Hashimoto, Wu,
  Desbordes, Rothschild, Phan, Wang, Nkinyili, Schick, Kornev,
  {Telleen-Lawton}, Tunduny, Gerstenberg, Chang, Neeraj, Khot, Shultz, Shaham,
  Misra, Demberg, Nyamai, Raunak, Ramasesh, Prabhu, Padmakumar, Srikumar,
  Fedus, Saunders, Zhang, Vossen, Ren, Tong, Zhao, Wu, Shen, Yaghoobzadeh,
  Lakretz, Song, Bahri, Choi, Yang, Hao, Chen, Belinkov, Hou, Hou, Bai, Seid,
  Zhao, Wang, Wang, Wang, and Wu}]{srivastava_2022_ImitationGameQuantifying}
Aarohi Srivastava, Abhinav Rastogi, Abhishek Rao, Abu Awal~Md Shoeb, Abubakar
  Abid, Adam Fisch, Adam~R. Brown, Adam Santoro, Aditya Gupta, Adri{\`a}
  {Garriga-Alonso}, Agnieszka Kluska, Aitor Lewkowycz, Akshat Agarwal, Alethea
  Power, Alex Ray, Alex Warstadt, Alexander~W. Kocurek, Ali Safaya, Ali Tazarv,
  Alice Xiang, Alicia Parrish, Allen Nie, Aman Hussain, Amanda Askell, Amanda
  Dsouza, Ambrose Slone, Ameet Rahane, Anantharaman~S. Iyer, Anders Andreassen,
  Andrea Madotto, Andrea Santilli, Andreas Stuhlm{\"u}ller, Andrew Dai, Andrew
  La, Andrew Lampinen, Andy Zou, Angela Jiang, Angelica Chen, Anh Vuong,
  Animesh Gupta, Anna Gottardi, Antonio Norelli, Anu Venkatesh, Arash
  Gholamidavoodi, Arfa Tabassum, Arul Menezes, Arun Kirubarajan, Asher
  Mullokandov, Ashish Sabharwal, Austin Herrick, Avia Efrat, Aykut Erdem, Ayla
  Karaka{\c s}, B.~Ryan Roberts, Bao~Sheng Loe, Barret Zoph, Bart{\l}omiej
  Bojanowski, Batuhan {\"O}zyurt, Behnam Hedayatnia, Behnam Neyshabur, Benjamin
  Inden, Benno Stein, Berk Ekmekci, Bill~Yuchen Lin, Blake Howald, Cameron
  Diao, Cameron Dour, Catherine Stinson, Cedrick Argueta, C{\'e}sar~Ferri
  Ram{\'i}rez, Chandan Singh, Charles Rathkopf, Chenlin Meng, Chitta Baral,
  Chiyu Wu, Chris {Callison-Burch}, Chris Waites, Christian Voigt,
  Christopher~D. Manning, Christopher Potts, Cindy Ramirez, Clara~E. Rivera,
  Clemencia Siro, Colin Raffel, Courtney Ashcraft, Cristina Garbacea, Damien
  Sileo, Dan Garrette, Dan Hendrycks, Dan Kilman, Dan Roth, Daniel Freeman,
  Daniel Khashabi, Daniel Levy, Daniel~Mosegu{\'i} Gonz{\'a}lez, Danielle
  Perszyk, Danny Hernandez, Danqi Chen, Daphne Ippolito, Dar Gilboa, David
  Dohan, David Drakard, David Jurgens, Debajyoti Datta, Deep Ganguli, Denis
  Emelin, Denis Kleyko, Deniz Yuret, Derek Chen, Derek Tam, Dieuwke Hupkes,
  Diganta Misra, Dilyar Buzan, Dimitri~Coelho Mollo, Diyi Yang, Dong-Ho Lee,
  Ekaterina Shutova, Ekin~Dogus Cubuk, Elad Segal, Eleanor Hagerman, Elizabeth
  Barnes, Elizabeth Donoway, Ellie Pavlick, Emanuele Rodola, Emma Lam, Eric
  Chu, Eric Tang, Erkut Erdem, Ernie Chang, Ethan~A. Chi, Ethan Dyer, Ethan
  Jerzak, Ethan Kim, Eunice~Engefu Manyasi, Evgenii Zheltonozhskii, Fanyue Xia,
  Fatemeh Siar, Fernando {Mart{\'i}nez-Plumed}, Francesca Happ{\'e}, Francois
  Chollet, Frieda Rong, Gaurav Mishra, Genta~Indra Winata, Gerard {de Melo},
  Germ{\'a}n Kruszewski, Giambattista Parascandolo, Giorgio Mariani, Gloria
  Wang, Gonzalo {Jaimovitch-L{\'o}pez}, Gregor Betz, Guy {Gur-Ari}, Hana
  Galijasevic, Hannah Kim, Hannah Rashkin, Hannaneh Hajishirzi, Harsh Mehta,
  Hayden Bogar, Henry Shevlin, Hinrich Sch{\"u}tze, Hiromu Yakura, Hongming
  Zhang, Hugh~Mee Wong, Ian Ng, Isaac Noble, Jaap Jumelet, Jack Geissinger,
  Jackson Kernion, Jacob Hilton, Jaehoon Lee, Jaime~Fern{\'a}ndez Fisac,
  James~B. Simon, James Koppel, James Zheng, James Zou, Jan Koco{\'n}, Jana
  Thompson, Jared Kaplan, Jarema Radom, Jascha {Sohl-Dickstein}, Jason Phang,
  Jason Wei, Jason Yosinski, Jekaterina Novikova, Jelle Bosscher, Jennifer
  Marsh, Jeremy Kim, Jeroen Taal, Jesse Engel, Jesujoba Alabi, Jiacheng Xu,
  Jiaming Song, Jillian Tang, Joan Waweru, John Burden, John Miller, John~U.
  Balis, Jonathan Berant, J{\"o}rg Frohberg, Jos Rozen, Jose
  {Hernandez-Orallo}, Joseph Boudeman, Joseph Jones, Joshua~B. Tenenbaum,
  Joshua~S. Rule, Joyce Chua, Kamil Kanclerz, Karen Livescu, Karl Krauth,
  Karthik Gopalakrishnan, Katerina Ignatyeva, Katja Markert, Kaustubh~D. Dhole,
  Kevin Gimpel, Kevin Omondi, Kory Mathewson, Kristen Chiafullo, Ksenia
  Shkaruta, Kumar Shridhar, Kyle McDonell, Kyle Richardson, Laria Reynolds, Leo
  Gao, Li~Zhang, Liam Dugan, Lianhui Qin, Lidia {Contreras-Ochando},
  Louis-Philippe Morency, Luca Moschella, Lucas Lam, Lucy Noble, Ludwig
  Schmidt, Luheng He, Luis~Oliveros Col{\'o}n, Luke Metz, L{\"u}tfi~Kerem {\c
  S}enel, Maarten Bosma, Maarten Sap, Maartje {ter Hoeve}, Maheen Farooqi,
  Manaal Faruqui, Mantas Mazeika, Marco Baturan, Marco Marelli, Marco Maru,
  Maria Jose~Ram{\'i}rez Quintana, Marie Tolkiehn, Mario Giulianelli, Martha
  Lewis, Martin Potthast, Matthew~L. Leavitt, Matthias Hagen, M{\'a}ty{\'a}s
  Schubert, Medina~Orduna Baitemirova, Melody Arnaud, Melvin McElrath,
  Michael~A. Yee, Michael Cohen, Michael Gu, Michael Ivanitskiy, Michael
  Starritt, Michael Strube, Micha{\l} Sw{\k{e}}drowski, Michele Bevilacqua,
  Michihiro Yasunaga, Mihir Kale, Mike Cain, Mimee Xu, Mirac Suzgun, Mo~Tiwari,
  Mohit Bansal, Moin Aminnaseri, Mor Geva, Mozhdeh Gheini, Mukund~Varma T,
  Nanyun Peng, Nathan Chi, Nayeon Lee, Neta Gur-Ari Krakover, Nicholas Cameron,
  Nicholas Roberts, Nick Doiron, Nikita Nangia, Niklas Deckers, Niklas
  Muennighoff, Nitish~Shirish Keskar, Niveditha~S. Iyer, Noah Constant, Noah
  Fiedel, Nuan Wen, Oliver Zhang, Omar Agha, Omar Elbaghdadi, Omer Levy, Owain
  Evans, Pablo Antonio~Moreno Casares, Parth Doshi, Pascale Fung, Paul~Pu
  Liang, Paul Vicol, Pegah Alipoormolabashi, Peiyuan Liao, Percy Liang, Peter
  Chang, Peter Eckersley, Phu~Mon Htut, Pinyu Hwang, Piotr Mi{\l}kowski, Piyush
  Patil, Pouya Pezeshkpour, Priti Oli, Qiaozhu Mei, Qing Lyu, Qinlang Chen,
  Rabin Banjade, Rachel~Etta Rudolph, Raefer Gabriel, Rahel Habacker,
  Ram{\'o}n~Risco Delgado, Rapha{\"e}l Milli{\`e}re, Rhythm Garg, Richard
  Barnes, Rif~A. Saurous, Riku Arakawa, Robbe Raymaekers, Robert Frank, Rohan
  Sikand, Roman Novak, Roman Sitelew, Ronan LeBras, Rosanne Liu, Rowan Jacobs,
  Rui Zhang, Ruslan Salakhutdinov, Ryan Chi, Ryan Lee, Ryan Stovall, Ryan
  Teehan, Rylan Yang, Sahib Singh, Saif~M. Mohammad, Sajant Anand, Sam
  Dillavou, Sam Shleifer, Sam Wiseman, Samuel Gruetter, Samuel~R. Bowman,
  Samuel~S. Schoenholz, Sanghyun Han, Sanjeev Kwatra, Sarah~A. Rous, Sarik
  Ghazarian, Sayan Ghosh, Sean Casey, Sebastian Bischoff, Sebastian Gehrmann,
  Sebastian Schuster, Sepideh Sadeghi, Shadi Hamdan, Sharon Zhou, Shashank
  Srivastava, Sherry Shi, Shikhar Singh, Shima Asaadi, Shixiang~Shane Gu, Shubh
  Pachchigar, Shubham Toshniwal, Shyam Upadhyay, Shyamolima, Debnath, Siamak
  Shakeri, Simon Thormeyer, Simone Melzi, Siva Reddy, Sneha~Priscilla Makini,
  Soo-Hwan Lee, Spencer Torene, Sriharsha Hatwar, Stanislas Dehaene, Stefan
  Divic, Stefano Ermon, Stella Biderman, Stephanie Lin, Stephen Prasad,
  Steven~T. Piantadosi, Stuart~M. Shieber, Summer Misherghi, Svetlana
  Kiritchenko, Swaroop Mishra, Tal Linzen, Tal Schuster, Tao Li, Tao Yu, Tariq
  Ali, Tatsu Hashimoto, Te-Lin Wu, Th{\'e}o Desbordes, Theodore Rothschild,
  Thomas Phan, Tianle Wang, Tiberius Nkinyili, Timo Schick, Timofei Kornev,
  Timothy {Telleen-Lawton}, Titus Tunduny, Tobias Gerstenberg, Trenton Chang,
  Trishala Neeraj, Tushar Khot, Tyler Shultz, Uri Shaham, Vedant Misra, Vera
  Demberg, Victoria Nyamai, Vikas Raunak, Vinay Ramasesh, Vinay~Uday Prabhu,
  Vishakh Padmakumar, Vivek Srikumar, William Fedus, William Saunders, William
  Zhang, Wout Vossen, Xiang Ren, Xiaoyu Tong, Xinran Zhao, Xinyi Wu, Xudong
  Shen, Yadollah Yaghoobzadeh, Yair Lakretz, Yangqiu Song, Yasaman Bahri, Yejin
  Choi, Yichi Yang, Yiding Hao, Yifu Chen, Yonatan Belinkov, Yu~Hou, Yufang
  Hou, Yuntao Bai, Zachary Seid, Zhuoye Zhao, Zijian Wang, Zijie~J. Wang, Zirui
  Wang, and Ziyi Wu. 2022.
\newblock \href {https://doi.org/10.48550/arXiv.2206.04615} {Beyond the
  {{Imitation Game}}: {{Quantifying}} and extrapolating the capabilities of
  language models}.

\bibitem[{Taylor et~al.(2022)Taylor, Kardas, Cucurull, Scialom, Hartshorn,
  Saravia, Poulton, Kerkez, and Stojnic}]{taylor_2022_GalacticaLargeLanguage}
Ross Taylor, Marcin Kardas, Guillem Cucurull, Thomas Scialom, Anthony
  Hartshorn, Elvis Saravia, Andrew Poulton, Viktor Kerkez, and Robert Stojnic.
  2022.
\newblock \href {https://doi.org/10.48550/arXiv.2211.09085} {Galactica: {{A
  Large Language Model}} for {{Science}}}.

\bibitem[{Thoppilan et~al.(2022)Thoppilan, De~Freitas, Hall, Shazeer,
  Kulshreshtha, Cheng, Jin, Bos, Baker, Du, Li, Lee, Zheng, Ghafouri, Menegali,
  Huang, Krikun, Lepikhin, Qin, Chen, Xu, Chen, Roberts, Bosma, Zhao, Zhou,
  Chang, Krivokon, Rusch, Pickett, Srinivasan, Man, {Meier-Hellstern}, Morris,
  Doshi, Santos, Duke, Soraker, Zevenbergen, Prabhakaran, Diaz, Hutchinson,
  Olson, Molina, {Hoffman-John}, Lee, Aroyo, Rajakumar, Butryna, Lamm, Kuzmina,
  Fenton, Cohen, Bernstein, Kurzweil, {Aguera-Arcas}, Cui, Croak, Chi, and
  Le}]{thoppilan_2022_LaMDALanguageModels}
Romal Thoppilan, Daniel De~Freitas, Jamie Hall, Noam Shazeer, Apoorv
  Kulshreshtha, Heng-Tze Cheng, Alicia Jin, Taylor Bos, Leslie Baker, Yu~Du,
  YaGuang Li, Hongrae Lee, Huaixiu~Steven Zheng, Amin Ghafouri, Marcelo
  Menegali, Yanping Huang, Maxim Krikun, Dmitry Lepikhin, James Qin, Dehao
  Chen, Yuanzhong Xu, Zhifeng Chen, Adam Roberts, Maarten Bosma, Vincent Zhao,
  Yanqi Zhou, Chung-Ching Chang, Igor Krivokon, Will Rusch, Marc Pickett,
  Pranesh Srinivasan, Laichee Man, Kathleen {Meier-Hellstern}, Meredith~Ringel
  Morris, Tulsee Doshi, Renelito~Delos Santos, Toju Duke, Johnny Soraker, Ben
  Zevenbergen, Vinodkumar Prabhakaran, Mark Diaz, Ben Hutchinson, Kristen
  Olson, Alejandra Molina, Erin {Hoffman-John}, Josh Lee, Lora Aroyo, Ravi
  Rajakumar, Alena Butryna, Matthew Lamm, Viktoriya Kuzmina, Joe Fenton, Aaron
  Cohen, Rachel Bernstein, Ray Kurzweil, Blaise {Aguera-Arcas}, Claire Cui,
  Marian Croak, Ed~Chi, and Quoc Le. 2022.
\newblock \href {https://doi.org/10.48550/arXiv.2201.08239} {{{LaMDA}}:
  {{Language Models}} for {{Dialog Applications}}}.

\bibitem[{Touvron et~al.(2023{\natexlab{a}})Touvron, Lavril, Izacard, Martinet,
  Lachaux, Lacroix, Rozi{\`e}re, Goyal, Hambro, Azhar, Rodriguez, Joulin,
  Grave, and Lample}]{touvron_2023_LLaMAOpenEfficient}
Hugo Touvron, Thibaut Lavril, Gautier Izacard, Xavier Martinet, Marie-Anne
  Lachaux, Timoth{\'e}e Lacroix, Baptiste Rozi{\`e}re, Naman Goyal, Eric
  Hambro, Faisal Azhar, Aurelien Rodriguez, Armand Joulin, Edouard Grave, and
  Guillaume Lample. 2023{\natexlab{a}}.
\newblock \href {https://doi.org/10.48550/arXiv.2302.13971} {{{LLaMA}}:
  {{Open}} and {{Efficient Foundation Language Models}}}.

\bibitem[{Touvron et~al.(2023{\natexlab{b}})Touvron, Martin, Stone, Albert,
  Almahairi, Babaei, Bashlykov, Batra, Bhargava, Bhosale, Bikel, Blecher,
  Ferrer, Chen, Cucurull, Esiobu, Fernandes, Fu, Fu, Fuller, Gao, Goswami,
  Goyal, Hartshorn, Hosseini, Hou, Inan, Kardas, Kerkez, Khabsa, Kloumann,
  Korenev, Koura, Lachaux, Lavril, Lee, Liskovich, Lu, Mao, Martinet, Mihaylov,
  Mishra, Molybog, Nie, Poulton, Reizenstein, Rungta, Saladi, Schelten, Silva,
  Smith, Subramanian, Tan, Tang, Taylor, Williams, Kuan, Xu, Yan, Zarov, Zhang,
  Fan, Kambadur, Narang, Rodriguez, Stojnic, Edunov, and
  Scialom}]{touvron_2023_LlamaOpenFoundation}
Hugo Touvron, Louis Martin, Kevin Stone, Peter Albert, Amjad Almahairi, Yasmine
  Babaei, Nikolay Bashlykov, Soumya Batra, Prajjwal Bhargava, Shruti Bhosale,
  Dan Bikel, Lukas Blecher, Cristian~Canton Ferrer, Moya Chen, Guillem
  Cucurull, David Esiobu, Jude Fernandes, Jeremy Fu, Wenyin Fu, Brian Fuller,
  Cynthia Gao, Vedanuj Goswami, Naman Goyal, Anthony Hartshorn, Saghar
  Hosseini, Rui Hou, Hakan Inan, Marcin Kardas, Viktor Kerkez, Madian Khabsa,
  Isabel Kloumann, Artem Korenev, Punit~Singh Koura, Marie-Anne Lachaux,
  Thibaut Lavril, Jenya Lee, Diana Liskovich, Yinghai Lu, Yuning Mao, Xavier
  Martinet, Todor Mihaylov, Pushkar Mishra, Igor Molybog, Yixin Nie, Andrew
  Poulton, Jeremy Reizenstein, Rashi Rungta, Kalyan Saladi, Alan Schelten, Ruan
  Silva, Eric~Michael Smith, Ranjan Subramanian, Xiaoqing~Ellen Tan, Binh Tang,
  Ross Taylor, Adina Williams, Jian~Xiang Kuan, Puxin Xu, Zheng Yan, Iliyan
  Zarov, Yuchen Zhang, Angela Fan, Melanie Kambadur, Sharan Narang, Aurelien
  Rodriguez, Robert Stojnic, Sergey Edunov, and Thomas Scialom.
  2023{\natexlab{b}}.
\newblock \href {https://doi.org/10.48550/arXiv.2307.09288} {Llama 2: {{Open
  Foundation}} and {{Fine-Tuned Chat Models}}}.

\bibitem[{Wei et~al.(2022)Wei, Tay, Bommasani, Raffel, Zoph, Borgeaud,
  Yogatama, Bosma, Zhou, Metzler, Chi, Hashimoto, Vinyals, Liang, Dean, and
  Fedus}]{wei_2022_EmergentAbilitiesLarge}
Jason Wei, Yi~Tay, Rishi Bommasani, Colin Raffel, Barret Zoph, Sebastian
  Borgeaud, Dani Yogatama, Maarten Bosma, Denny Zhou, Donald Metzler, Ed~H.
  Chi, Tatsunori Hashimoto, Oriol Vinyals, Percy Liang, Jeff Dean, and William
  Fedus. 2022.
\newblock \href {https://openreview.net/forum?id=yzkSU5zdwD} {Emergent
  {{Abilities}} of {{Large Language Models}}}.
\newblock \emph{Transactions on Machine Learning Research}.

\end{thebibliography}
\bibliographystyle{acl_natbib}

\end{document}